\documentclass[11pt,a4paper]{article}
\usepackage[hyperref]{acl2019}
\usepackage{times}
\usepackage{latexsym}

\usepackage{url}

\usepackage{times}
\usepackage{amsmath}
\usepackage{amsfonts}
\usepackage{amssymb}
\usepackage{latexsym}
\usepackage{booktabs}
\usepackage{color} 
\usepackage{setspace}

\usepackage{multirow}

\usepackage{url}
\usepackage{graphicx} 
\usepackage{subfigure}
\aclfinalcopy 

\title{Dual Co-Matching Network for Multi-choice Reading Comprehension}

\author{Shuailiang Zhang$^{1,2,3}$,Hai Zhao$^{1,2,3,}$\thanks{$\ $ Corresponding author. This paper was partially supported by National Key Research and Development Program of China (No. 2017YFB0304100), 
		Key Projects of National Natural Science Foundation of China (U1836222 and 61733011), 
		Key Project of National Society Science Foundation of China (No. 15-ZDA041), 
		The Art and Science Interdisciplinary Funds of Shanghai Jiao Tong University (No. 14JCRZ04)}$\ $, Yuwei Wu$^{1,2,3}$, Zhuosheng Zhang$^{1,2,3}$, 
	\\  \Large\textbf{Xi Zhou$^{4}$, Xiang Zhou$^{4}$}\\
	$^{1}$Department of Computer Science and Engineering, Shanghai Jiao Tong University \\
	$^{2}$Key Laboratory of Shanghai Education Commission for Intelligent Interaction \\ and Cognitive Engineering, Shanghai Jiao Tong University, Shanghai, China\\
	$^{3}$MoE Key Lab of Artificial Intelligence, AI Institute, Shanghai Jiao Tong University, Shanghai, China\\
	$^{4}$CloudWalk Technology, Shanghai, China\\
	{\tt \{zsl123\}@sjtu.edu.cn, zhaohai@cs.sjtu.edu.cn,}\\
	{\tt  \{will8821,zhangzs\}@sjtu.edu.cn, \{zhouxi,zhouxiang\}@cloudwalk.cn}\\
}
\date{}

\begin{document}
	\maketitle
	\begin{abstract}
		Multi-choice reading comprehension is a challenging task to select an answer from a set of candidates when given passage and question.  This work proposes dual co-matching network which models the relationship among passage, question and answer bidirectionally. The experimental results on RACE, ROCStories and COIN Shared Task 1 show that our model obtains state-of-the-art results and even the single model outperforms the human performance on RACE dataset.
	\end{abstract}
	
	\section{Introduction}
	
	Machine reading comprehension (MRC) is a fundamental and long-standing goal of natural language understanding which aims to teach the machine to answer a question automatically according to a given passage \cite{Hermann15, Rajpurkar-D16, NguyenRSGTMD16}. 
	
	In this paper, we focus on multi-choice MRC task such as ROCStories \cite{roc2016-corpus}, COIN Shared Task 1 \cite{semeval11} and RACE \cite{Lai-2017} which requests to choose the right answer from a set of candidate ones according to given passage and question. Different from MRC datasets such as SQuAD \cite{Rajpurkar-D16} and NewsQA \cite{Trischler-W17} where the expected answer is usually in the form of a short span from the given passage, answer in multi-choice MRC is non-extractive and may not appear in the original passage, which allows rich types of questions such as commonsense reasoning and passage summarization, as illustrated by two example questions in Table \ref{table1}. In Q1, the model suffers the distraction from the similarity between \textit{different routes} in answer A and \textit{many routes} in the passage and probably selects answer A by mistake. Thus multi-choice MRC is more challenging and requires a more in-depth understanding of the given passage and question \cite{Lai-2017, Khashabi-N18-1023}.  
	
	To well handle the multi-choice MRC problem, a common solution should carefully model the relationship among the triplet of three sequences, passage (\textbf{P}), question (\textbf{Q}) and answer (\textbf{A}) with a matching module to determine the answer. Previous matching strategies \cite{Wang-2018, tang2019multi, Chen2018ConvolutionalSA, ocn} are usually unidirectional which only calculate question-aware passage representation and ignore passage-aware question representation when modeling the relationship between passage and question. These unidirectional matching methods obviously cannot take the best of information between two sequences. Besides, previous methods usually fail to cover all the relationship among the triplet. For example, \cite{Wang-2018} ignore the relationship between \textbf{Q} and \textbf{A}. Unlike other MRC tasks such as SQuAD that focus on modeling the relationship between \textbf{P} and \textbf{Q}, the three sequences \textbf{P}, \textbf{Q} and \textbf{A} are equally important and any pairwise relationship among the triplet in multi-choice MRC should be sufficiently under consideration.

	In this work, we propose dual co-matching network (DMN) which incorporates all the pairwise relationships among the \{\textbf{P}, \textbf{Q}, \textbf{A}\} triplet bidirectionally. In detail, we model the passage-question, passage-answer and question-answer pairwise relationship simultaneously and bidirectionally for each triplet. What's more, we exploit the gated mechanism to fuse the representations from two directions which is proved to be more effective than a simple concatenation strategy.        
	
	\begin{table}[t!]
		\begin{center}
			\begin{tabular}{|p{6.9cm}|}
				\hline
				\textbf{Passage}: Runners in a relay race pass a stick in one direction. However, merchants passed silk, gold, fruit, and glass along the Silk Road in more than one direction. They earned their living by traveling the famous Silk Road. ... \textbf{The Silk Road was made up of many routes, not one smooth path.} They passed through what are now 18 countries. The routes crossed mountains and deserts and had many dangers of hot sun, deep snow and even battles... \\ 
				\hline
				\textbf{Question}: The Silk Road became less important because \_ .\\    
				\quad A. it was made up of different routes\\  
				\quad B. silk trading became less popular\\
				\quad \textbf{C. sea travel provided easier routes}  \\ 
				\quad D. people needed fewer foreign goods \\
				\hline
			\end{tabular}
		\end{center}
		\caption{\label{table1} An example passage with related question and options from RACE dataset. The ground-truth answer and the evidence sentences in the passage are in \textbf{bold}.}
	\end{table}
	
	Besides, our model leverages the latest pre-trained resources BERT \cite{Devlin-18} by utilizing its output as our contextual embedding, which plays a role of encoder. Our model is evaluated on the multi-choice MRC benchmark challenge, RACE and ROCStories, which reports new state-of-the-art by 2.6\% and 1.2\% better than previous best results on RACE and ROCStories, respectively. Especially, our single model performs even better than human turkers on the RACE full dataset, which is the first milestone achievement ever since the RACE challenge has been set up.
	
	\begin{figure*}[t!]
		\centering
		\includegraphics[width=5in]{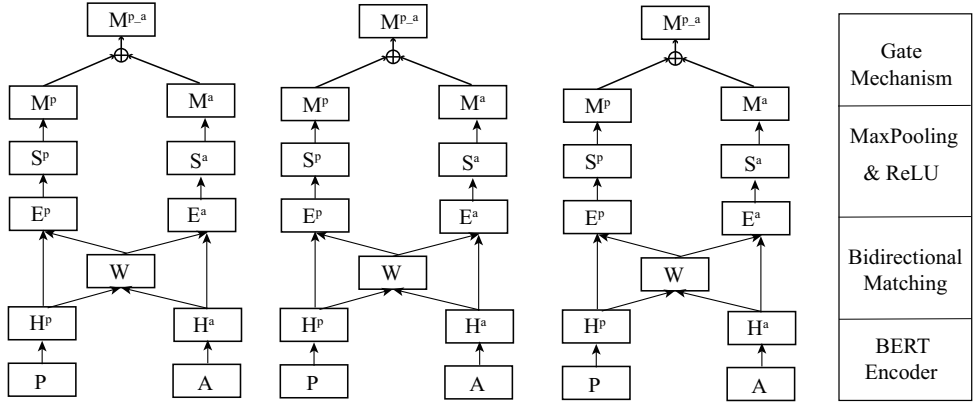}
		
		\caption{The framework of our model. P-Passage, Q-Question, A-Answer. $\oplus$ indicates the gated mechanism in Eq. \ref{eq3}.}
		\label{figure1}
	\end{figure*}
	
	\section{Model}
	
	For the task of multi-choice reading comprehension, the machine is given a passage (\textbf{P}), a question (\textbf{Q}), and a set of candidate answers (\textbf{A}) and the goal is to select the correct answer from the candidates. 
	
	\subsection{Encoder} \label{Encoding} 
	This layer encodes each token in passage and question into a fixed-length vector following BERT \cite{Devlin-18}. The passage, the question, and the candidate answer are encoded as follows:
	\begin{equation} 
	\textbf{H}^p=Bert(\textbf{P}),\textbf{H}^q=Bert(\textbf{Q}),\textbf{H}^a=Bert(\textbf{A}) \nonumber
	\end{equation}
	where $Bert$ represents the BERT model which returns the last layer output in BERT. $\textbf{H}^p \in R^{|P| \times l}$, $\textbf{H}^q \in R^{|Q| \times l}$, and $\textbf{H}^a \in R^{|A| \times l}$ are sequence representation of the passage, question and answer, respectively.
	$|P|$, $|Q|$, $|A|$ are the sequence length of the passage, the question and the candidate answer, respectively. $l$ is the dimension of the hidden state.
	
	\subsection{Matching} \label{Matching} 
	
	To fully model the relationship in a \{\textbf{P}, \textbf{Q}, \textbf{A}\} triplet, We make use of the bidirectional matching strategy and gated mechanism to get all pairwise matching representation among the triplet, including passage-answer, passage-question and question-answer representation. This part shows how to build matching representation for passage-answer sequence pair and it is the same for the other two pairs.
	
	\textbf{Bidirectional matching} representation between the passage $\textbf{H}^p$ and answer $\textbf{H}^a$ can be calculated as follows:
	\begin{equation} \label{eq2}
	\begin{split}
	\textbf{G}^{qa}&=SoftMax(\textbf{H}^pW{\textbf{H}^{a}}^T), \\
	\textbf{E}^{p}&=\textbf{G}^{qa}\textbf{H}^{a},
	\textbf{E}^{a}={\textbf{G}^{qa}}^T\textbf{H}^{p},\\
	\textbf{S}^{p} &= ReLU(\textbf{E}^{p}W_1),\\
	\textbf{S}^{a} &= ReLU(\textbf{E}^{a}W_2),
	\end{split}
	\end{equation}
	where $W$, $W_1$, $W_2 \in R^{l \times l}$ are three learnable parameters. $\textbf{G}^{qa} \in R^{|P| \times |A|}$ is the attention weight matrix between the passage and the answer. $\textbf{E}^{p} \in R^{|P| \times l}, \textbf{E}^{a} \in R^{|A| \times l}$ represent answer-aware passage representation and passage-aware answer representation, respectively. \cite{Wang-2018, tang2019multi} use elementwise subtraction and multiplication to fuse $\textbf{E}^{p}$ and $\textbf{H}^{p}$ (\emph{i.e.}, $[\textbf{E}^{p} \ominus \textbf{H}^{p}; \textbf{E}^{p} \otimes \textbf{H}^{p}]$) which is has shown not good enough as such processing breaks the symmetry of equation. Symmetric representations from both directions show essentially helpful for our bidirectional architecture.
	
	\textbf{Gated mechanism} is used to fuse $\textbf{S}^{p}$ and $\textbf{S}^{a}$ because gate is more powerful to regulate the information flow \cite{highway}. The final results also prove that the gated mechanism works better than the concatenation strategy. The final representation of passage-answer pair is calculated as follows:
	\begin{equation} \label{eq3}
	\begin{split}
	\textbf{M}^{p} &= MaxPooling(\textbf{S}^{p}),\\
	\textbf{M}^{a} &= MaxPooling(\textbf{S}^{a}),\\
	g & = \sigma(\textbf{M}^{p}W_3 + \textbf{M}^{a}W_4 + b),\\
	\textbf{M}^{p\_a} &= g * \textbf{M}^{p}+(1-g) * \textbf{M}^{a},
	\end{split}
	\end{equation} 
	where $W_3, W_4 \in R^{l \times l}$ and $b \in R^{l}$ are three learnable parameters.  After a row-wise max pooling operation, we get the aggregation representation $\textbf{M}^{p} \in R^{l}$ and $\textbf{M}^{a} \in R^{l}$. $g \in R^l$ is the reset gate. $\textbf{M}^{p\_a} \in R^{l}$ is the final bidirectional matching representation of the passage-answer sequence pair. 
	
	Passage-question and question-answer sequence matching representation $\textbf{M}^{p\_q}, \textbf{M}^{q\_a} \in R^{l}$ can be calculated in the same procedure using Eq. (\ref{eq2}) and (\ref{eq3}). The framework of our model is shown in Figure \ref{figure1}.

	\subsection{Objective function}
	In matching layer, we build the matching representation $\textbf{M}^{p\_q}, \textbf{M}^{p\_a}, \textbf{M}^{q\_a}$ for three sequence pairs. Finally, we concatenate them as the final representation $\textbf{C} \in R^{3l}$ for each passage-question-answer triplet. We can build $\textbf{C}_i$ for each \{$P, Q, A_i$\} triplet. So the objective function can be computed as follows:
	\begin{equation} \label{eq4}
	\begin{split}
	\textbf{C} &= [\textbf{M}^{p\_q}; \textbf{M}^{p\_a}; \textbf{M}^{q\_a}],\\
	L(A_i|P,Q) &= -log{\frac{exp(V^T\textbf{C}_i)}{\sum_{j=1}^4{exp(V^T\textbf{C}_j)}}}, 
	\end{split}
	\end{equation}
	where $V \in R^l$ is a learnable parameter. 
	
	\section{Experiment}
	We evaluate our models on Large-scale ReAding Comprehension Dataset From Examinations (RACE) dataset \cite{Lai-2017}, which consists of two subsets: RACE-M and RACE-H corresponding to middle school and high school difficulty level. RACE contains 27,933 passages and 97,687 questions in total, which is recognized as one of the largest and most difficult datasets in multi-choice reading comprehension. Besides, we also evaluate our model on the ROCStories (Spring 2016) dataset which collects 50k five-sentence commonsense stories. One correct ending sentence needs to be selected from two options when given four sentences in ROCStories dataset.  
	
	\subsection{Experiment Setup} 
	We use BERT as our encoder. The max sequence length is set to 512. A dropout rate of 0.1 is applied to every BERT layer. We optimize the model using BertAdam \cite{Devlin-18} optimizer with a learning rate 5e-6. In the matching layer, the dropout rate is set to 0.3. We train for 10 epochs with batch size 4 when BERT$_{large}$ is used as encoder.  
	
	\subsection{Comparison with Baselines} 
	Performance comparison against all baseline models is shown in Table \ref{table2}. Turkers is the performance of Amazon Turkers on a random subset of the RACE test set. Ceiling is the percentage of unambiguous questions in the test set. From the table, we can see that the large-scale pre-trained language models OpenAI GPT \cite{Radford2018ImprovingLU} and BERT \cite{Devlin-18} surpass all previous models only by fine-tuning.
	
	\begin{table}[t!]
		\begin{center}
			\small
			\resizebox{\linewidth}{!}{
				\begin{tabular*}{\hsize}{@{}@{\extracolsep{\fill}}llllll@{}}
					\hline
					\bf Model         &RACE-M &RACE-H &RACE\\
					\hline 
					DFN \cite{Xu-2017} &	51.5 &45.7 &47.4\\
					HAF \cite{zhuhaichao2018hierarchical} &45.0 &46.4 &46.0\\
					MRU \cite{Tay-2018} &57.7 &47.4 &50.4\\
					HCM \cite{Wang-2018} &55.8 &48.2 &50.4\\
					MMN \cite{tang2019multi} &61.1 &52.2 &54.7\\
					GPT \cite{Radford2018ImprovingLU} &62.9 &57.4 &59.0\\
					RSM \cite{SunKai-2018}&69.2 &61.5 &63.8\\
					
					BERT$_{base}$   &71.1&62.3 &65.0 \\
					BERT$_{large}$  &76.6&70.1 &72.0 \\
					OCN \cite{ocn}$^*$  &78.4&71.5 &73.5 \\
					\hline
					Our Models\\
					DMN$_{base}$   &72.3 &64.2 &66.5\\
					DMN$_{large}$    &77.6 &70.1 &72.3\\
					DMN$^*_{large}$    &\bf79.5 &\bf71.8 &\bf74.1\\
					\hline
					Human Performance \\
					Turkers     &85.1&69.4 &\underline{73.3}\\
					Ceiling    &95.4&94.2 &94.5\\
					\hline
			\end{tabular*}}
		\end{center}
		\caption{\label{table2} Experiment results on RACE test set. $\dagger$ means it is statistically significant to the models ablating either the bidirectional matching or gated mechanism. $^*$ indicates ensemble model. DMN$_{base}$ uses BERT$_{base}$ as encoder and DMN$_{large}$ uses BERT$_{large}$ as encoder.}
	\end{table}

	RSM \cite{SunKai-2018} and OCN \cite{ocn} further improve the results based on GPT and BERT respectively. RSM \cite{SunKai-2018} introduces three domain-independent reading strategies and OCN \cite{ocn} compares candidate answers at word-level to better identify their correlations. In this work, we propose the bidirectional matching strategy and gated mechanism to model the pairwise sequence relationship among the passage-question-answer triplet. 
	
	Comparison results show that our model is powerful and even the sing model outperforms all baselines and achieves new state-of-the-art accuracy. Our ensemble model further improves the performance for 1.5\%. More specifically, our single model is the first to outperform Amazon Turkers on RACE full dataset as shown in Table \ref{table2}, which is one milestone achievement ever since the RACE challenge has been set up.
	
	Comparison with previous models on ROCStories is shown in Table \ref{table3}. Our model also outperforms all the baselines and achieves state-of-the-art accuracy by improving for 1.2\% compared to BERT$_{large}$.
	
	\subsection{Ablation Study} 
	In this work, we mainly focus on show two core model improvements, (1) the bidirectional matching strategy and (2) the gated mechanism. Ablation experiments on RACE are shown in Table \ref{table2}. We observe 1.5\% performance decrease by only using unidirectional matching. In detail, we only build answer-aware passage representation without considering passage-aware answer representation when modeling the passage-answer sequence pair relationship (\emph{i.e.}, only use $\textbf{S}^p$ as the matching representation without $\textbf{S}^a$ in Eq. (\ref{eq2})). We also observe 0.5\% decrease by replacing the gated mechanism with directly concatenating two sequences (\emph{i.e.}, concatenate $\textbf{M}^p$ and $\textbf{M}^a$ as $\textbf{M}^{p\_a}$ in Eq.(\ref{eq3})). The ablation experiments on ROCStories dataset further prove above descriptions. 
	
	What's more, the relationship between question-answer sequence pair is not under sufficient consideration in previous work such as \cite{Wang-2018}. So we remove the question-answer matching from the model (\emph{i.e.}, only concatenate $\textbf{M}^{p\_q}$ and $\textbf{M}^{p\_a}$ as $\textbf{C}$ in Eq. (\ref{eq4})) which leads to 0.4\% decrease, indicating that this relationship should get more attention. 
	
	\begin{table}[t!]
		\begin{center}
			\small
			\resizebox{\linewidth}{!}{
				\begin{tabular*}{\hsize}{@{}@{\extracolsep{\fill}}llcl@{}}
					\hline
					&Approach  &Accuracy\\
					\hline 
					ROCStories &OFT \cite{Radford2018ImprovingLU} & 86.5\\
					&RSM \cite{SunKai-2018} &88.3\\
					&BERT &90.2	\\
					&DMN+BERT (Ours) &\textbf{91.4} \\
					\hline
					COIN Task 1 &Multi-Finetune BERT &84.2\\
					&BERTFT &73.1\\
					&Knowledge Graph Fusion BERT &80.7\\
					&DMN+BERT (Ours) &\textbf{86.8}\\
					&BERT+XLNet &90.5\\
					&Multi-Finetune-XLnet &90.6\\
					&DMN+XLNet (Ours)&\textbf{91.1}$^*$\\
					\hline
			\end{tabular*}}
		\end{center}
		\caption{\label{table3} Performance comparison on the ROCStories (Spring 2016) and COIN test set. $^*$: this result is evaluated on the COIN dev set because we missed the deadline of the evaluation period on COIN test set.}
	\end{table}
	
	
	\section{Conclusions}
	In this work, we propose Dual Co-Matching Network (DMN) to model the bidirectional sequence relationship among the passage, question, and the candidate answer. By incorporating the well pre-trained BERT in an innovative way, our model achieves new state-of-the-art in ROCStories and RACE benchmarks, outperforming the previous state-of-the-art model by 2.6\% in RACE and 1.2\% in ROCStories. In addition, for the first time the single model outperforms human turkers on the RACE full dataset.

\bibliography{acl2019}

\begin{thebibliography}{17}
\expandafter\ifx\csname natexlab\endcsname\relax\def\natexlab#1{#1}\fi

\bibitem[{Chen et~al.(2018)Chen, Cui, Ma, and Wang}]{Chen2018ConvolutionalSA}
Zhipeng Chen, Yiming Cui, Wentao Ma, and Shijin Wang. 2018.
\newblock {Convolutional Spatial Attention Model for Reading Comprehension with
  Multiple-Choice Questions}.
\newblock volume abs/1811.08610.

\bibitem[{Devlin et~al.(2018)Devlin, Chang, Lee, and Toutanova}]{Devlin-18}
Jacob Devlin, Ming{-}Wei Chang, Kenton Lee, and Kristina Toutanova. 2018.
\newblock {BERT: Pre-training of Deep Bidirectional Transformers for Language
  Understanding}.
\newblock \emph{CoRR}, abs/1810.04805.

\bibitem[{Hermann et~al.(2015)Hermann, Kocisk{\'{y}}, Grefenstette, Espeholt,
  Kay, Suleyman, and Blunsom}]{Hermann15}
Karl~Moritz Hermann, Tom{\'{a}}s Kocisk{\'{y}}, Edward Grefenstette, Lasse
  Espeholt, Will Kay, Mustafa Suleyman, and Phil Blunsom. 2015.
\newblock {Teaching Machines to Read and Comprehend}.
\newblock \emph{CoRR}, abs/1506.03340.

\bibitem[{Khashabi et~al.(2018)Khashabi, Chaturvedi, Roth, Upadhyay, and
  Roth}]{Khashabi-N18-1023}
Daniel Khashabi, Snigdha Chaturvedi, Michael Roth, Shyam Upadhyay, and Dan
  Roth. 2018.
\newblock {Looking Beyond the Surface: A Challenge Set for Reading
  Comprehension over Multiple Sentences }.
\newblock In \emph{Proceedings of the 2018 Conference of the North American
  Chapter of the Association for Computational Linguistics: Human Language
  Technologies, Volume 1 (Long Papers)}, pages 252--262. Association for
  Computational Linguistics.

\bibitem[{Lai et~al.(2017)Lai, Xie, Liu, Yang, and Hovy}]{Lai-2017}
Guokun Lai, Qizhe Xie, Hanxiao Liu, Yiming Yang, and Eduard Hovy. 2017.
\newblock {RACE: Large-scale ReAding Comprehension Dataset From Examinations}.
\newblock In \emph{Proceedings of the 2017 Conference on Empirical Methods in
  Natural Language Processing}, pages 785--794. Association for Computational
  Linguistics.

\bibitem[{Nguyen et~al.(2016)Nguyen, Rosenberg, Song, Gao, Tiwary, Majumder,
  and Deng}]{NguyenRSGTMD16}
Tri Nguyen, Mir Rosenberg, Xia Song, Jianfeng Gao, Saurabh Tiwary, Rangan
  Majumder, and Li~Deng. 2016.
\newblock {{MS} {MARCO:} {A} Human Generated MAchine Reading COmprehension
  Dataset}.
\newblock \emph{CoRR}, abs/1611.09268.

\bibitem[{Radford(2018)}]{Radford2018ImprovingLU}
Alec Radford. 2018.
\newblock {Improving Language Understanding by Generative Pre-Training}.

\bibitem[{Rajpurkar et~al.(2016)Rajpurkar, Zhang, Lopyrev, and
  Liang}]{Rajpurkar-D16}
Pranav Rajpurkar, Jian Zhang, Konstantin Lopyrev, and Percy Liang. 2016.
\newblock {SQuAD: 100,000+ Questions for Machine Comprehension of Text}.
\newblock In \emph{Proceedings of the 2016 Conference on Empirical Methods in
  Natural Language Processing}, pages 2383--2392. Association for Computational
  Linguistics.

\bibitem[{Ran et~al.(2019)Ran, Li, Hu, and Zhou}]{ocn}
Qiu Ran, Peng Li, Weiwei Hu, and Jie Zhou. 2019.
\newblock Option comparison network for multiple-choice reading comprehension.
\newblock \emph{CoRR}, abs/1903.03033.

\bibitem[{Srivastava et~al.(2015)Srivastava, Greff, and Schmidhuber}]{highway}
Rupesh~Kumar Srivastava, Klaus Greff, and J{\"{u}}rgen Schmidhuber. 2015.
\newblock Highway networks.
\newblock \emph{CoRR}, abs/1505.00387.

\bibitem[{Sun et~al.(2018)Sun, Yu, Yu, and Cardie}]{SunKai-2018}
Kai Sun, Dian Yu, Dong Yu, and Claire Cardie. 2018.
\newblock {Improving Machine Reading Comprehension with General Reading
  Strategies}.
\newblock \emph{CoRR}, abs/1810.13441.

\bibitem[{Tang et~al.(2019)Tang, Cai, and Zhuo}]{tang2019multi}
Min Tang, Jiaran Cai, and Hankz~Hankui Zhuo. 2019.
\newblock {Multi-Matching Network for Multiple Choice Reading Comprehension}.
\newblock In \emph{Thirty-Second AAAI Conference on Artificial Intelligence}.

\bibitem[{Tay et~al.(2018)Tay, Tuan, and Hui}]{Tay-2018}
Yi~Tay, Luu~Anh Tuan, and Siu~Cheung Hui. 2018.
\newblock {Multi-range Reasoning for Machine Comprehension}.
\newblock \emph{CoRR}, abs/1803.09074.

\bibitem[{Trischler et~al.(2017)Trischler, Wang, Yuan, Harris, Sordoni,
  Bachman, and Suleman}]{Trischler-W17}
Adam Trischler, Tong Wang, Xingdi Yuan, Justin Harris, Alessandro Sordoni,
  Philip Bachman, and Kaheer Suleman. 2017.
\newblock {NewsQA: A Machine Comprehension Dataset}.
\newblock In \emph{Proceedings of the 2nd Workshop on Representation Learning
  for NLP}, pages 191--200. Association for Computational Linguistics.

\bibitem[{Wang et~al.(2018)Wang, Yu, Jiang, and Chang}]{Wang-2018}
Shuohang Wang, Mo~Yu, Jing Jiang, and Shiyu Chang. 2018.
\newblock {A Co-Matching Model for Multi-choice Reading Comprehension}.
\newblock In \emph{Proceedings of the 56th Annual Meeting of the Association
  for Computational Linguistics (Volume 2: Short Papers)}, pages 746--751.
  Association for Computational Linguistics.

\bibitem[{Xu et~al.(2017)Xu, Liu, Gao, Shen, and Liu}]{Xu-2017}
Yichong Xu, Jingjing Liu, Jianfeng Gao, Yelong Shen, and Xiaodong Liu. 2017.
\newblock {Towards Human-level Machine Reading Comprehension: Reasoning and
  Inference with Multiple Strategies}.
\newblock \emph{CoRR}, abs/1711.04964.

\bibitem[{Zhu et~al.(2018)Zhu, Wei, Qin, and Liu}]{zhuhaichao2018hierarchical}
Haichao Zhu, Furu Wei, Bing Qin, and Ting Liu. 2018.
\newblock {Hierarchical Attention Flow for Multiple-choice Reading
  Comprehension}.
\newblock In \emph{Thirty-Second AAAI Conference on Artificial Intelligence}.

\end{thebibliography}
\bibliographystyle{acl_natbib}

\appendix

\end{document}